\documentclass[conference]{IEEEtran}
\IEEEoverridecommandlockouts

\usepackage{cite}
\usepackage{amsmath,amssymb,amsfonts}
\usepackage{algorithmic}
\usepackage{graphicx}
\usepackage{textcomp}
\usepackage[table]{xcolor}
\usepackage{booktabs}
\usepackage{marvosym}
\usepackage{fancyhdr}
\usepackage[colorlinks,
            linkcolor=blue,
            anchorcolor=blue,
            citecolor=blue]{hyperref}

\def\BibTeX{{\rm B\kern-.05em{\sc i\kern-.025em b}\kern-.08em
    T\kern-.1667em\lower.7ex\hbox{E}\kern-.125emX}}

\begin{document}

\title{% UFO-DETR: An Novel Detector for Frequency-Aware UAV Tiny-Object Detection
    UFO-DETR: Frequency-Guided End-to-End Detector for UAV Tiny Objects
}

\author{
    Yuankai Chen\textsuperscript{1,*}, 
    Kai Lin\textsuperscript{1,2,*}, 
    Qihong Wu\textsuperscript{1}, 
    Xinxuan Yang\textsuperscript{1}, 
    Jiashuo Lai\textsuperscript{1}, 
    Ruoen Chen\textsuperscript{1}, \\
    Haonan Shi\textsuperscript{1}, 
    Minfan He\textsuperscript{3} and
    Meihua Wang\textsuperscript{1,\Letter} \\
    \textsuperscript{1}Easy Lab, College of Mathematics and Informatics, South China Agricultural University, Guangzhou, China\\
    \textsuperscript{2}School of Computer Science and Engineering, Southeast University, Nanjing, China\\
    \textsuperscript{3}School of Mathematics, Foshan University, Foshan, China \\
    \{ab9853211, scau\_kk, wqhwqh, yangxx, shelly, 15768209196, shihaonan\}@stu.scau.edu.cn \\
    heminfan1980@126.com, wangmeihua@scau.edu.cn \\
    {\small *~Equal Contribution, \Letter~Corresponding author}
}

\maketitle

% 页眉
\thispagestyle{fancy}
\chead{Proceedings of the 2026 29th International Conference on Computer Supported Cooperative Work in Design}
\renewcommand{\headrulewidth}{0pt}

\begin{abstract}
Small target detection in UAV imagery faces significant challenges such as scale variations, dense distribution, and the dominance of small targets. Existing algorithms rely on manually designed components, and general-purpose detectors are not optimized for UAV images, making it difficult to balance accuracy and complexity. To address these challenges, this paper proposes an end-to-end object detection framework, UFO-DETR, which integrates an LSKNet-based backbone network to optimize the receptive field and reduce the number of parameters. By combining the DAttention and AIFI modules, the model flexibly models multi-scale spatial relationships, improving multi-scale target detection performance. Additionally, the DynFreq-C3 module is proposed to enhance small target detection capability through cross-space frequency feature enhancement. Experimental results show that, compared to RT-DETR-L, the proposed method offers significant advantages in both detection performance and computational efficiency, providing an efficient solution for UAV edge computing.
\end{abstract}


\section{Introduction}
\IEEEPARstart{U}{nmanned} aerial vehicles (UAVs), due to their compact size, maneuverability, and ease of operation~\cite{[1]}, are widely used in tasks like rescuing personnel and inspecting power lines, where accurate and efficient object detection is crucial. However, high-altitude imaging and dynamic viewpoints often result in targets occupying a small fraction of pixels and exhibiting significant scale variations, while the wide field of view introduces complex and diverse backgrounds. These factors make small-object detection particularly challenging, with achieving a balance between detection performance and computational efficiency becoming a key research issue in UAV-based detection.

In response to the above challenges, end-to-end object detection methods, which eliminate NMS post-processing and thereby simplify the detection pipeline, have attracted considerable attention. However, existing approaches still suffer from several bottlenecks, in particular the excessive computational overhead of the networks~\cite{[2]} and their limited capability to extract features for small objects~\cite{[3]}, which prevents them from achieving real-time and high-accuracy small-object detection on resource-constrained UAV platforms. To address such challenges, this paper proposes a lightweight small-object detection model, \textbf{U}AV Tiny-Object \textbf{F}requency-\textbf{O}ptimized \textbf{DE}tection \textbf{TR}ansformer (\textbf{UFO-DETR}), built upon the RT-DETR framework, with the goal of reducing computational complexity while preserving high-precision recognition of small objects, thereby providing a new perspective that jointly considers efficiency and accuracy for UAV-oriented small-object detection.

The main contributions of this paper are summarized as follows:
\begin{itemize}
\item We propose UFO-DETR, a real-time end-to-end object detector for UAV imagery that achieves high accuracy while alleviating the challenge of balancing performance and computational complexity.
\item We adopt Large Selective Kernel Network (LSKNet) as the backbone; its dynamic receptive field design reduces model parameters and improves inference speed.
\item We embed deformable attention into AIFI to dynamically adjust sampling locations on feature maps, thereby modeling multi-scale spatial relations and improving feature capture for objects of different sizes.
\item We design DynFreq-C3, a cross spatial-frequency module, fuses spatial features in the frequency domain to recover high-frequency details, enhancing small-object detection in complex backgrounds.
\end{itemize}

\begin{figure*}[t]
    \centering
    \includegraphics[width=0.8\textwidth]{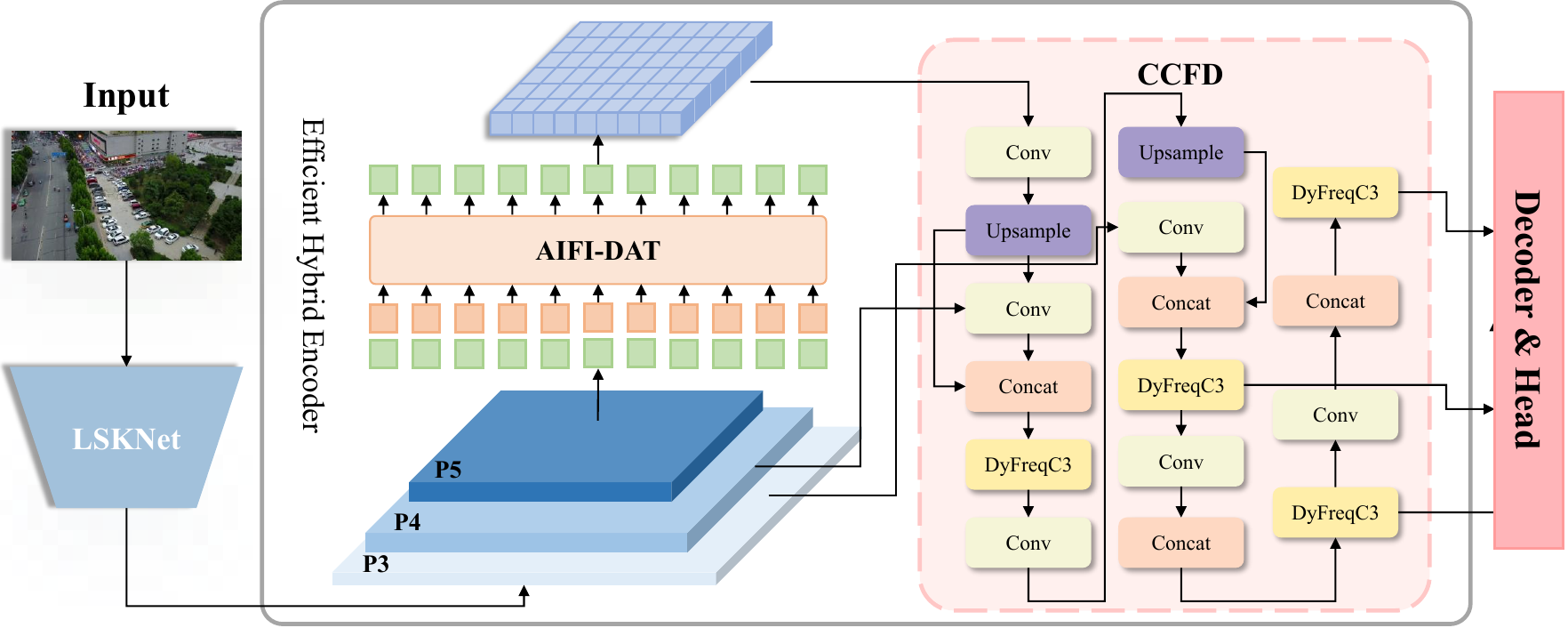}
    \caption{Overview of the UFO-DETR. It is divided into the backbone, encoder, and detection head. The backbone is rebuilt using LSKNet. The decoder is implemented through CCFD based on DyFreqC3 design and AIFI with integrated deformable attention. The detection head consists of the decoder and auxiliary prediction heads, completing the final object detection task.}
    \vspace{-10pt}
    \label{fig.1}
\end{figure*}

\section{RELATED WORK}
\subsection{Object Detection in UAV Imagery}
Object detection in UAV imagery faces several challenges, including small object sizes, drastic scale variations, and complex backgrounds, which significantly degrade the accuracy and robustness of conventional detectors in this scenario. Approaches such as FS-SSD and CSIPN introduce contextual information and enhance local regions to improve the detection accuracy of small objects under complex backgrounds~\cite{[4],[5]}. Another line of work, including PBR-YOLO and Piglet-YOLO, applies channel or spatial attention to reweight multi-scale features, suppress background responses, and enhance the discriminative capability for small objects~\cite{[6],[7],[8]}. SGGF-Net incorporates Gaussian priors to optimize positive/negative sample assignment for small objects and combines global context enhancement with attention-guided multi-scale feature fusion, thereby improving the localization accuracy and robustness of small objects in complex scenes~\cite{[9]}. However, on resource-constrained UAV platforms, achieving a balance between model compactness and high detection performance remains challenging, and most methods rely on spatial-domain features and fail to fully leverage frequency-domain texture information.

\subsection{Real-time detection of UAV images}
With the advances in deep learning, existing object detectors can be roughly categorized into two families. The first consists of two-stage detectors based on region proposals (e.g., R-CNN, Faster R-CNN)~\cite{[10],[11]}, which typically achieve high accuracy but are computationally expensive and thus difficult to deploy under the real-time requirements of UAV platforms. The second comprises one-stage detectors, among which the YOLO-family methods are representative~\cite{[12],[13],[14]}. Although they offer high inference speed, they still rely on NMS as a post-processing step, and their accuracy for small-object detection in complex scenes remains limited.

In contrast, the Transformer-based end-to-end detector RT-DETR (Real-Time Detection Transformer)~\cite{[15]} leverages efficient multi-scale feature fusion and a dynamic query mechanism to completely remove anchor design and NMS post-processing, achieving a more favorable trade-off between efficiency and performance for UAV small-object detection.

\subsection{Multi-Scale Feature Fusion}
Multi-scale feature fusion, which combines local details with global semantics, is widely used to improve small-object detection performance~\cite{[16]}. Existing methods can be roughly divided into two groups. The first enlarges the receptive field using multi-scale or dilated convolutions to enhance contextual modeling; RFB~\cite{[17]} employs multi-branch convolutions to simulate diverse receptive fields, and ASPP~\cite{[18]} fuses multi-rate dilated convolutions with global pooling, but the former is still limited in representing small objects, while the latter incurs high computational redundancy and struggles to satisfy real-time requirements. The second group builds feature-pyramid-based cross-scale fusion structures to combine high-level semantics and low-level details; PANet~\cite{[19]} introduces a top-down pyramid with bottom-up path aggregation, and BiFPN~\cite{[20]} adopts learnable weights for efficient multi-scale fusion, yet these architectures remain relatively complex and provide only limited gains for small-object detection.

In view of the limitations of these methods in terms of detection performance and computational efficiency, we build on LSKNet, integrating large-kernel convolutions with deformable attention to optimize contextual modeling and multi-scale feature enhancement, improving real-time small-object detection.

\section{PROPOSED METHOD}

To address the challenges of small target features lack sufficient saliency in multi-scale scenarios and complex backgrounds, this paper proposes a lightweight detection framework, UFO-DETR. The framework is designed for UAV imagery and aims to balance detection performance and the constrained computational resources on edge devices. Based on RT-DETR, this paper introduces LSKNet as the backbone network, effectively reducing both the number of parameters and computational cost through the Large Kernel Selection and Spatial Selection mechanisms. Subsequently, the DAttention module is introduced, which dynamically generates sampling points to adaptively focus on key regions in the feature map, addressing the feature differences across targets of different scales. Finally, this paper proposes the DynFreq-C3 module integrates spatial and frequency domain information, enhancing the model’s ability to detect small targets by exploiting texture details and contextual associations. The overall framework of the model and key components are shown in Fig.~\ref{fig.1}.

\subsection{LSKNet Backbone Network}
\begin{figure}[h]
    \vspace{-10pt}
    \centering
    \includegraphics[width=0.4\linewidth]{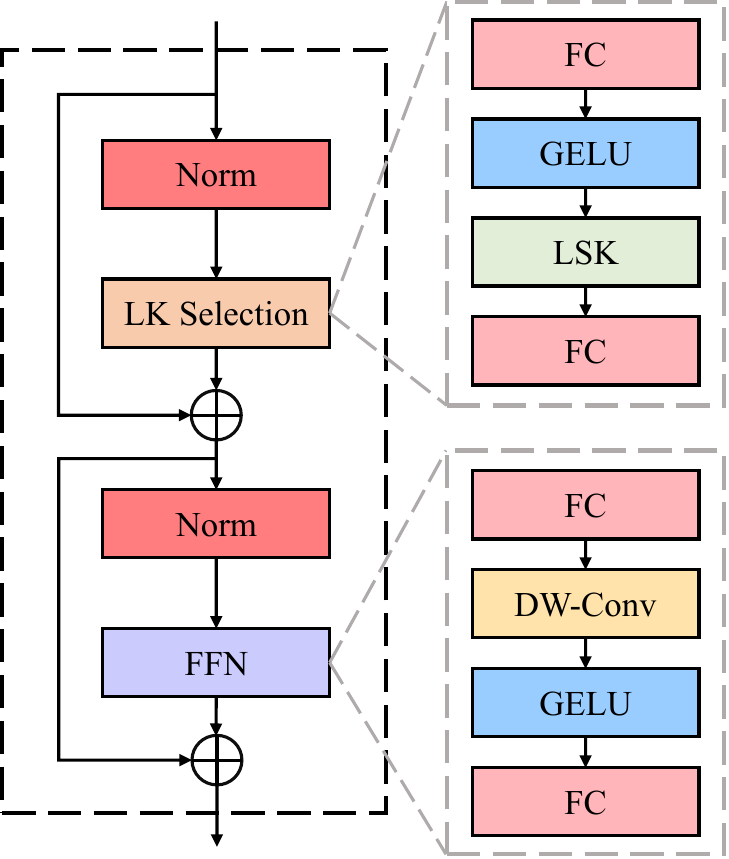}
    \caption{Structure of the LSKNet Block and LK Selection.}
    \vspace{-5pt}
    \label{fig.2}
\end{figure}

RT-DETR employs ResNet-based backbone networks, which perform excellently in feature extraction. However, their high computational cost makes it challenging to achieve real-time operation on resource constrained UAV platforms.At the same time, its fixed convolution kernel size and limited receptive field make it difficult to handle small targets with large scale variations and complex backgrounds in aerial imagery.The Large Selective Kernel Network proposed by Li et al.~\cite{[21]} is lightweight and can adaptively adjust its spatial receptive field, allowing it to more accurately capture small targets and the key information surrounding them. To achieve model lightweighting and effectively extract contextual information, this paper adopts LSKNet as the backbone network, with the structure of the LSKNet Block shown in Fig.~\ref{fig.2}.

The LSKNet backbone is composed of repeated LSKNet Blocks, each containing two residual sub-modules: the LK Selection (Large Kernel Selection) and the FFN (Feed-Forward Network). The LK Selection module decomposes large convolutional kernels into smaller ones and, combined with the spatial selection mechanism, adaptively adjusts the receptive field to capture small-target details and context. This decomposition is achieved through depthwise separable convolutions and dilation adjustments, while the spatial selection mechanism dynamically assigns weights to regions via spatial pooling at different scales. This enables the model to focus on areas most relevant to the target. The FFN sub-module, which performs channel mixing and feature optimization, consists of fully connected layers, depthwise convolutions, GELU activations, and another fully connected layer, further enhancing feature representation.

The workflow of LSKNet can be represented by the following formula:

\vspace{-15pt}
\begin{equation}
Y = F_{\text{FFN}}(F_{\text{LK-Selection}}(X))
\label{eq.1}
\end{equation}
\vspace{-13pt}

Where $X$ represents the input feature map, $F_{\text{LK Selection}}$ denotes the processing of the input feature map through the Large Kernel Selection mechanism, and $F_{\text{FFN}}$ represents the processing by the Feed-Forward Network.

Therefore, LSKNet achieves dynamic spatial receptive field adjustment while maintaining a low computational cost, effectively improving detection accuracy and efficiency in small target scenarios with significant scale variations and complex backgrounds.

\subsection{Deformable Attention}
In UAV aerial imagery, the AIFI module focuses on extracting deep-layer feature maps, which leads to poor performance in small target recognition and complex backgrounds. Additionally, due to the limitations of the fixed positional encoding in its self-attention mechanism, it struggles to flexibly model spatial relationships at different scales, resulting in the loss of important features~\cite{[22]}.

Additionally, due to the variations in UAV viewpoints, challenges such as significant intra-class scale inconsistencies and target occlusion arise. To address these issues, this study introduces a deformable attention mechanism~\cite{[23]}. This mechanism dynamically adjusts the weight configuration of the feature map through learnable offsets, thereby enhancing the ability to capture features of targets at different scales.

\begin{figure}[h]
    \centering
    \includegraphics[width=\linewidth]{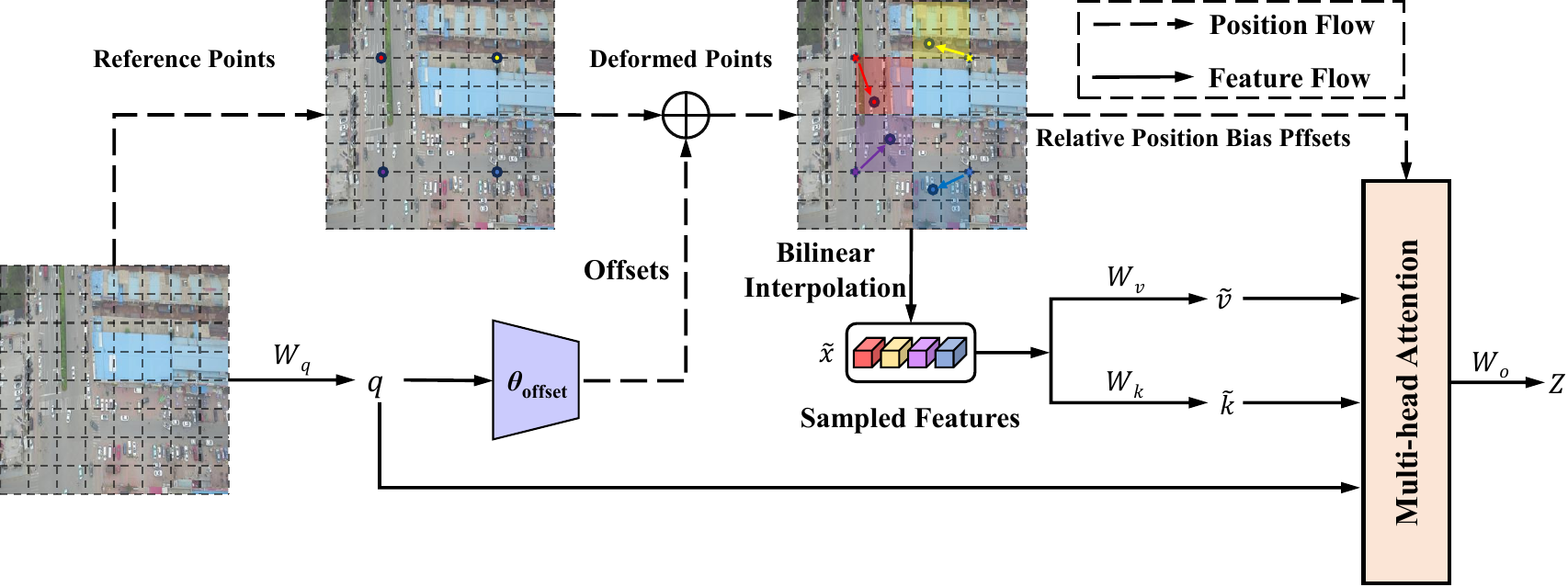}
    \caption{An illustration of Deformable Attention mechanism.}
    \vspace{-5pt}
    \label{fig.3}
\end{figure}

Fig.~\ref{fig.3} illustrates the working principle of the module. DAttention dynamically generates sampling points through an offset network, allowing it to adaptively focus on relevant regions in the feature map. Specifically, for the input feature map $x \in \mathbb{R}^{H \times W \times C}$ a uniformly spaced grid of reference points $p \in \mathbb{R}^{H_G \times W_G \times 2}$ is generated, and the reference points are normalized to the range $[-1, +1]$. Then, the feature map is queried by $q = x W_q$ and the offset network, consisting of operations and activation functions, calculates the offset $\Delta p = \theta_{\text{offset}}(q)$, dynamically adjusting the reference points. Afterward, the offset-adjusted reference points 
which dynamically adjusts the reference points. Afterward, the offset-adjusted $\hat{B}$ are sampled, obtaining the deformed feature map $\tilde{\chi}$. Finally, bilinear interpolation is used to perform weighted sampling on the deformed reference points, combined with the relative position offset matrix $R$.The final output feature $z(m)$ is then computed through the multi-head self-attention mechanism.

This process can be expressed as:

\begin{equation}
z(m) = \sigma\!\left(\frac{q(m)\,\tilde{k}(m)^{T}}{\sqrt{d}} + \varphi(\hat{B}; R)\right)\tilde{v}(m)
\label{eq.2}
\end{equation}

Where $\tilde{k}$ represents the Key vector; $\tilde{v}$ represents the Value vector; $d$ represents the dimension of the query vector; $m$ represents the $m_th$ vector; $\sigma$ represents the activation function, and $\varphi(\cdot)$ represents the position bias calculation function.

This study integrates the DAttention mechanism into the AIFI module, forming the DAttention-AIFI module. The module first performs a dimensional transformation operation on the 2D input feature $S$, converting it into a 1D vector. This 1D vector is then fed into the DAttention-AIFI module. Next, by combining multi-head self-attention (MHSA) with the feed-forward network (FFN), the input features are further enhanced. Finally, the output features of the module are converted back into a 2D format F, providing input support for the subsequent cross-scale feature fusion stage.

This process can be expressed as follows:

\begin{equation}
Q = K = V = \operatorname{Flatten}(S)
\label{eq.3}
\end{equation}

\vspace{-10pt}

\begin{equation}
F = \operatorname{Reshape}(\operatorname{DAttn}(Q, K, V))
\label{eq.4}
\end{equation}

Where Flatten represents the dimensional transformation operation; DAttn represents the deformable attention operation; Reshape represents the operation that reshapes the feature map back into the same shape as $S$.

By incorporating DAttention, the model can effectively address the intra-class scale inconsistency caused by UAV multi-scale viewpoints, thereby further enhancing the accuracy and robustness of object detection.

\subsection{DynFreq-C3}
\begin{figure}[h]
    \centering
    \includegraphics[width=0.6\linewidth]{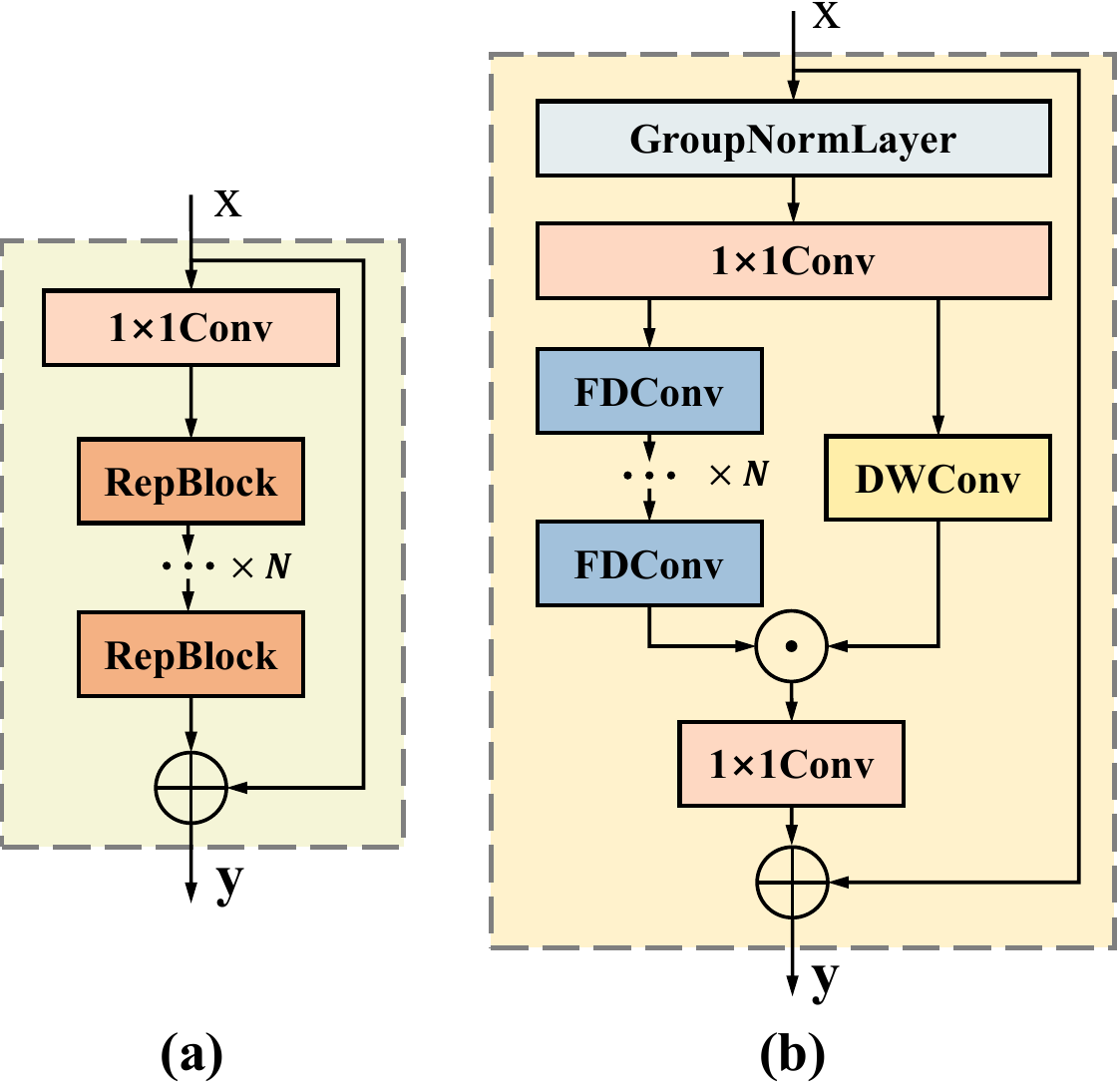}
    \caption{(a) RepC3 structure diagram; (b) DynFreq-C3 structure diagram.}
    \vspace{-15pt}
    \label{fig.4}
\end{figure}

To address challenges in UAV detection, where small targets are hard to distinguish from complex backgrounds, we propose DynFreq-C3 (Dynamic Frequency Convolution-based RepC3). As shown in Fig.~\ref{fig.4}, the original RepC3 module in RT-DETR consists of multiple RepConv modules, two branches, and a residual connection. Frequency domain information has a unique advantage in small target detection, as the edge and texture information of small targets is typically represented as high-frequency components in the frequency domain. Relying solely on spatial domain convolutions for feature fusion often leads to the loss of these high-frequency components. It uses FDConv~\cite{[24]} to capture high-frequency details, enhancing small target perception, while the DWConv branch improves spatial domain feature extraction, helping distinguish small targets from the background.

The computational process of the DynFreq-C3 is as shown in the formula:

\vspace{-10pt}
\begin{equation}
F_{0} = \operatorname{Conv}_{1\times1}(x)
\label{eq.5}
\end{equation}

\vspace{-10pt}

\begin{equation}
F_{\text{freq}} = \prod_{i=1}^{N} \operatorname{FDConv}(F_{i-1})
\label{eq.6}
\end{equation}

\vspace{-5pt}

\begin{equation}
F_{s} = \operatorname{DWConv}(F_{0})
\label{eq.7}
\end{equation}

\vspace{-10pt}

\begin{equation}
y = x + \operatorname{Conv}_{1\times1}(F_{\text{freq}} \cdot F_{s})
\label{eq.8}
\end{equation}

Where $x$ represents the original input feature map; $F_{i-1}$ represents the feature map after the $i-1$ convolution; $F_{\text{freq}}$ represents the feature map after $N\times\operatorname{FDConv}$ operations; $F_{s}$ represents the feature map processed through DWConv.

This paper designs the frequency domain information extraction method, which improves the recognition of small targets by extracting high-frequency information, effectively enhancing the model's ability to distinguish small target boundaries and details. Meanwhile, the depth in the spatial domain is finely divided and effectively reduced in computational complexity, making the model lighter.

\section{EXPERIMENT}
\subsection{Experimental Setup}
\textbf{Datasets:} In the experiment, we used the VisDrone2019 dataset~\cite{[25]} released by the AISKYEYE team at Tianjin University. The dataset contains 14,018 UAV images from different cities and traffic scenes, covering ten categories including pedestrians, vehicles, and bicycles. To meet the needs of this study, the dataset was split into a training set of 6,471 images, a validation set of 548 images, and a test set of 3,190 images. The challenges of this dataset include a large number of small-sized targets, complex occlusions, and uneven data distribution, which fully reflect the difficulties and challenges of object detection in UAV aerial imagery.

\textbf{Implementation Details:} All models were trained on a device with an NVIDIA GeForce RTX 3080ti (12GB VRAM) GPU. RT-DETR includes three variants: one with ResNet18, one with ResNet50, and one with EfficientFormerV2~\cite{[26]}. RT-DETR-L with EfficientFormerV2 is used as the baseline to build UFO-DETR. The training used 400 iterations with a batch size of 8, an initial learning rate of 0.01, and the SGD optimizer. For parameter optimization, AdamW was employed with a learning rate of 0.0001 and momentum of 0.9. Input images were resized to 640×640 pixels, and the Mosaic augmentation method was consistently applied. Performance was evaluated using six metrics: Precision (P), Recall (R), mean Average Precision (mAP), GFLOPs, and Model Size.

\begin{figure*}[t]
    \centering
    \includegraphics[width=\linewidth]{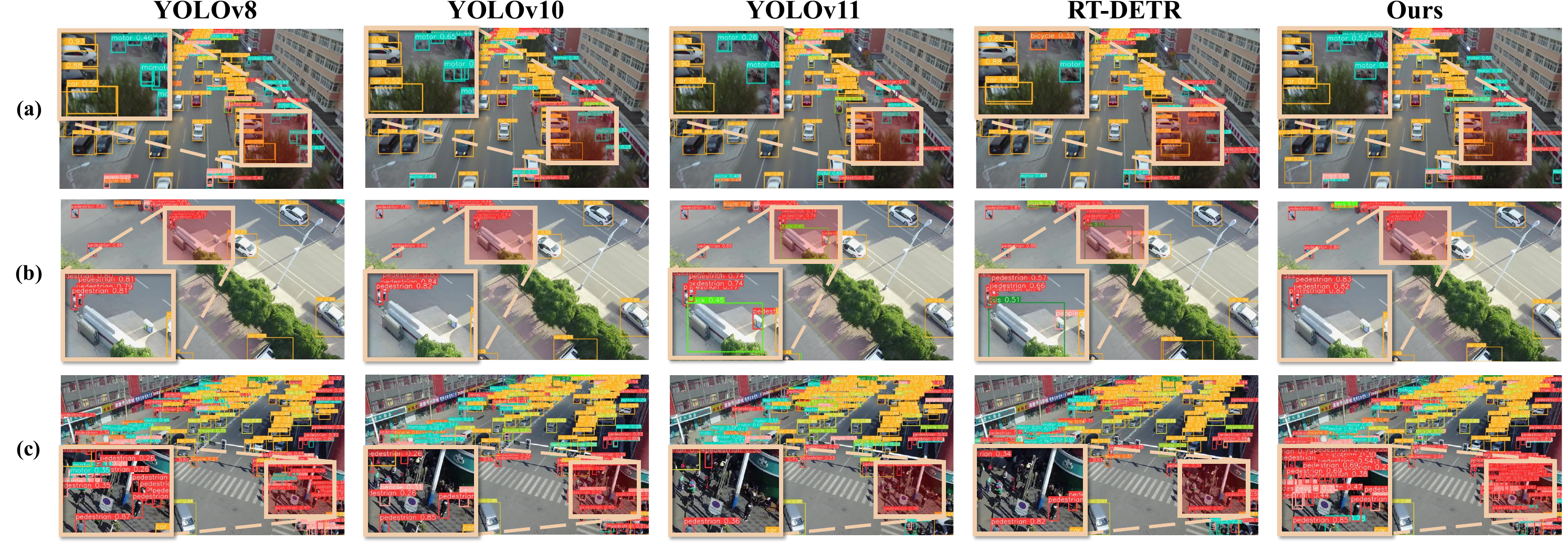}
    \caption{Visual comparison results of RTDETR-L, YOLOv8M, YOLOv10, YOLOv11M. (a) shows Repeated Detections of the same object, (b) corresponds to model false positives, and (c) denotes objects that the model failed to detect.}
    \vspace{-10pt}
    \label{fig.5}
\end{figure*}

\subsection{Ablation Study}
To validate the effectiveness of the various improvements in the experiment, we conducted ablation studies on the VisDrone2019 dataset to analyze the impact of each component on detection accuracy. Table 1 presents a performance comparison under different configurations, where the baseline represents the RT-DETR model, and A, B, and C represent the LSKNet backbone network, DAttention module, and DynFreq-C3 module. As shown in Table~\ref{tab.1}, the baseline model RT-DETR-R18 achieved a detection performance of 59.0\% Precision, 42.4\% Recall, and 43.5\% mAP50, but the model complexity is relatively high. After replacing the backbone network with LSKNet, the GFLOPs and Model Size were reduced to 37.6 and 26.0 MB, respectively, while the detection performance remained nearly unchanged. After adding the DAttention module to the AIFI module, the model's Recall and mAP50 were further improved to 43.0\% and 44.2\%, respectively. When all components were integrated, the UFO-DETR model achieved the highest detection performance, with 59.2\% Precision, 44.5\% Recall, and 46.1\% mAP50, demonstrating the cumulative effect of each module on detection accuracy. Additionally, the UFO-DETR model's 41.8 GFLOPs and 28.3 MB Model Size indicate that the proposed model is highly suitable for edge computing platforms with limited computational resources.

\vspace{-10pt}
\begin{table}[h]
    \caption{Results of the ablation study.}
    \centering
    \setlength{\tabcolsep}{3pt} % 调小列间距，避免过宽
    \small                     % 表格整体缩小一号字
    \resizebox{\columnwidth}{!}{% 保证表格宽度不超过当前栏宽
    \begin{tabular}{lccccc}
        \toprule
        \multicolumn{1}{c}{Model} & P/\% & R/\% & mAP50/\% & GFLOPs & Model Size/MB \\
        \midrule
        \raggedright baseline                 & 59.0 & 42.4 & 43.5 & 103.5 & 66.2 \\
        \raggedright baseline + A             & 58.1 & 41.7 & 43.4 & 37.6  & 26.0 \\
        \raggedright baseline + A + B         & 59.0 & 43.0 & 44.2 & 37.8  & 26.0 \\
        \raggedright baseline + A + B + C     & 59.2 & 44.5 & 46.1 & 41.8  & 28.3 \\
        \bottomrule
    \end{tabular}
    }
    \label{tab.1}
\end{table}

\subsection{Comparative experiments of different models}
To validate the effectiveness of the proposed method, this paper conducts a comprehensive comparison with mainstream object detection methods on the VisDrone2019 dataset. The experimental results, as shown in Table~\ref{tab.2}, reveal that among the YOLO series, the best-performing model, YOLOv8-m, achieved 53.5\% Precision, 37.4\% Recall, and 40.7\% mAP50. The RT-DETR series outperforms the YOLO series significantly on the detection task in this study, with even the underperforming RT-DETR-R18 surpassing YOLOv8-m in all performance metrics. However, while RT-DETR demonstrates strong performance, it comes at the cost of substantial computational resources, parameter size, and inference time. The proposed UFO-DETR model excels in both detection performance and computational efficiency. In terms of detection performance, our model improves Precision, Recall, and mAP50 by 0.2\%, 1.9\%, and 2.6\%, respectively, compared to RT-DETR-L. Additionally, the model achieves 41.8 GFLOPs and 28.3MB in Model Size, outperforming other mainstream models. The experimental results show that the proposed algorithm achieves an excellent balance between high accuracy and lightweight design in complex UAV aerial imagery environments.

\begin{table}[t]
    \centering
    \caption{Comparison Results with Other Object Detectors.}

    \setlength{\tabcolsep}{3pt}
    \small
    % 定义淡蓝色
    \definecolor{lightblue}{RGB}{244, 250, 252}
    \resizebox{\columnwidth}{!}{
    \begin{tabular}{lccccc}
        \toprule
        \textbf{Model} & \textbf{P/\%} & \textbf{R/\%} & \textbf{mAP50/\%} & \textbf{GFLOPs} & \textbf{Model Size/MB} \\
        \midrule
        YOLOv8-M     & 53.5 & 37.4 & 40.7 & 78.9  & 52   \\
        YOLOv8-L     & 51.8 & 39.7 & 40.6 & 164.9 & 87.7 \\
        YOLOv10-M    & 49.9 & 36.7 & 37.9 & 63.5  & 33.5 \\
        YOLOv10-L    & 51.6 & 38.4 & 39.8 & 126.4 & 52.2 \\
        YOLOv11-M    & 51.8 & 38.1 & 39.6 & 67.7  & 40.5 \\
        YOLOv11-L    & 53.0 & 38.3 & 39.8 & 86.6  & 51.2 \\
        RTDETR-L     & 59.0 & 42.4 & 43.5 & 103.5 & 66.2 \\
        RTDETR-R18   & 55.1 & 40.2 & 41.5 & 57.0  & 40.5 \\
        RTDETR-R50   & 58.3 & \textbf{44.5} & 45.2 & 129.6 & 86.1 \\
        \rowcolor{lightblue}
        Ours         & \textbf{59.2} & \textbf{44.5} & \textbf{46.1} & \textbf{41.8}  & \textbf{28.3} \\
        \bottomrule
    \end{tabular}
    \label{tab.2}
    }
\end{table}

\subsection{Visualization}
\begin{figure}[h]
    \centering
    \includegraphics[width=\linewidth]{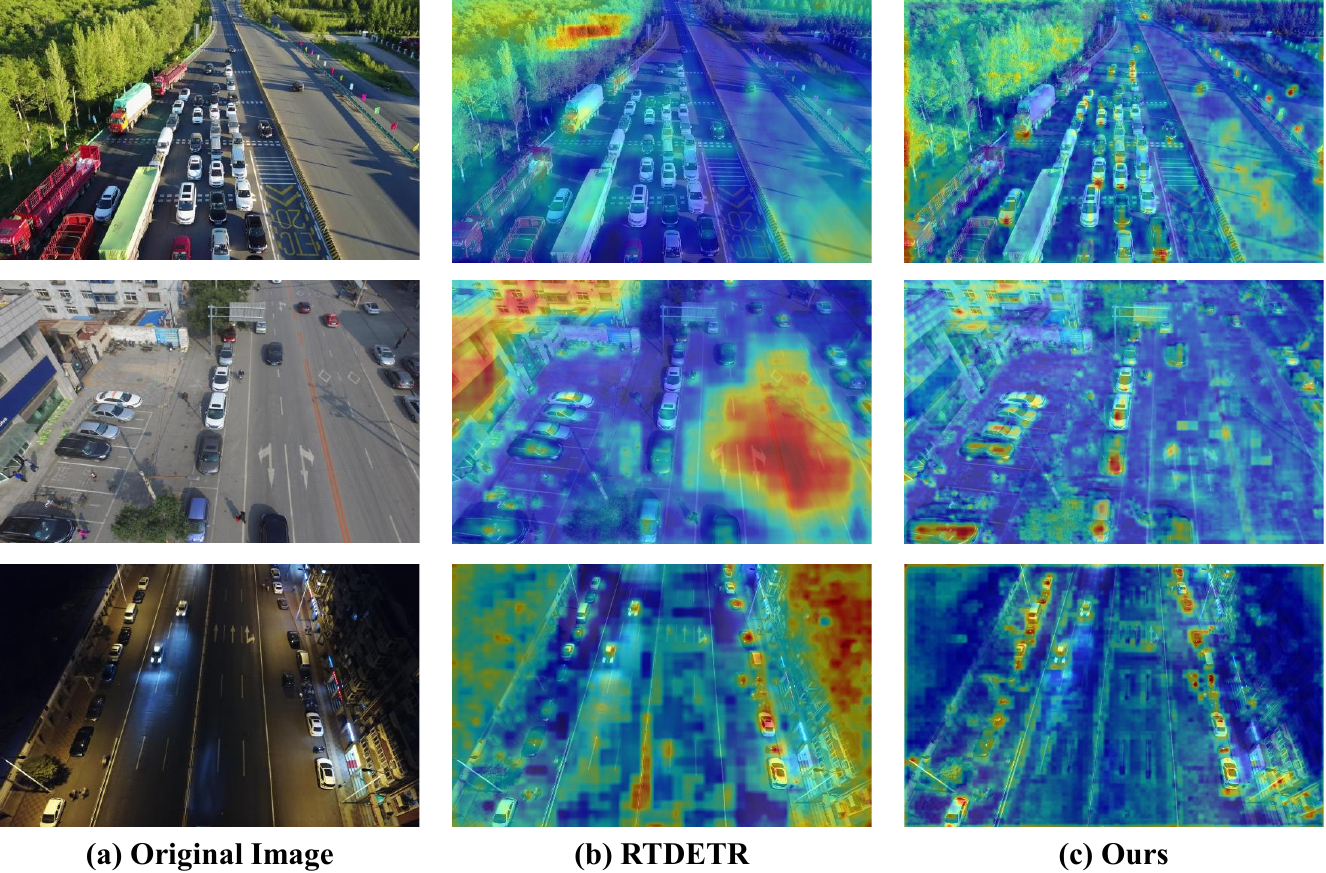}
    \caption{The heatmap of RTDETR-L and UFO-DETR. The brighter areas in the heatmap indicate stronger attention by the model.}
    \vspace{-10pt}
    \label{fig.6}
\end{figure}

To validate the performance of the proposed model, this paper qualitatively compares the detection results of UFO-DETR with several other detection models across various complex scenarios. As shown in Fig.~\ref{fig.5}, the proposed method achieves lower miss rates and false detection rates, thanks to the designed network model. In challenging scenarios, where multiple comparison models face issues such as duplicate detections, false detections, and missed detections, the UFO-DETR model performs well. Additionally, to visually demonstrate the model's ability to focus on detection targets, this paper uses Grad-CAM to generate activation heatmaps for different models, as shown in Fig.~\ref{fig.6}.

\section{CONCLUSION}
To address UAV multi-target detection challenges, we propose UFO-DETR, a real-time end-to-end detector designed for UAV imagery. This model balances accuracy and complexity, a common struggle for existing algorithms. Specifically, the LSKNet backbone network is introduced to replace the original model backbone, reducing model complexity while maintaining detection accuracy. Furthermore, the Cross-Spatial-Frequency module, DynFreq-C3, is designed for UAV images to capture both local and global contextual information of features, alleviating the interference of complex backgrounds and enhancing the recognition accuracy of small targets. Experimental results on the VisDrone2019 dataset demonstrate that UFO-DETR achieves higher accuracy than existing methods with lower computational cost while maintaining real-time inference speed. Future work will focus on addressing the redundant computational overhead caused by the positional relationship decoder within RT-DETR.

\section*{Acknowledgment}
This work is funded by the National Natural Science Foundation of China (61971052, 62476163), the Guangdong Provincial Key Areas Special Fund for General Higher Education Institutions (2024ZDZX4032), Guangdong Provincial Graduate Education Innovation Program Project (2025JGXM\_024) and National Undergraduate Innovation Training Program (202510564397).

% References
\bibliographystyle{IEEEtranBST2/IEEEtran}
\bibliography{references}

\end{document}